\documentclass[lettersize,journal]{IEEEtran}
\usepackage{amsmath,amsfonts}
\usepackage{algorithmic}
\usepackage{algorithm}
\usepackage{array}
\usepackage[caption=false,font=normalsize,labelfont=rm,textfont=rm]{subfig}
\usepackage{textcomp}
\usepackage{stfloats}
\usepackage{url}
\usepackage{verbatim}
\usepackage{graphicx}
\usepackage{cite}
\usepackage{mathrsfs}
\usepackage{booktabs}
\usepackage{multirow}
\begin{document}
\bstctlcite{IEEEexample:BSTcontrol} % make this the first directive after "\begin{document}"

\title{Cyber-Physical Steganography in\\Robotic Motion Control}

\author{Ching-Chun Chang, Yijie Lin and Isao Echizen
        % <-this % stops a space
\thanks{C.-C. Chang is with the Information and Society Research Division, National Institute of Informatics, Japan (email: ccchang@nii.ac.jp).}% <-this % stops a space
\thanks{Y. Lin is with the Department of Information Engineering and Computer Science, Feng Chia University, Taiwan.}% <-this % stops a space
\thanks{I. Echizen is with the Graduate School of Information Science and Technology, University of Tokyo, Japan.}% <-this % stops a space
\thanks{This work was supported in part by the Japan Society for the Promotion of Science (JSPS) under KAKENHI Grants (JP21H04907 and JP24H00732) and the Japan Science and Technology Agency (JST) under CREST Grants (JPMJCR18A6 and JPMJCR20D3) and AIP Acceleration Grants (JPMJCR24U3).}}

%\thanks{Manuscript received Month Day, Year; revised Month Day, Year.}}

\markboth{Chang \MakeLowercase{\textit{et al.}}: Cyber-Physical Steganography in Robotic Motion Control}%
{Chang \MakeLowercase{\textit{et al.}}: Cyber-Physical Steganography in Robotic Motion Control}

\maketitle

\begin{abstract}
Steganography, the art of information hiding, has continually evolved across visual, auditory and linguistic domains, adapting to the ceaseless interplay between steganographic concealment and steganalytic revelation. This study seeks to extend the horizons of what constitutes a viable steganographic medium by introducing a steganographic paradigm in robotic motion control. Based on the observation of the robot's inherent sensitivity to changes in its environment, we propose a methodology to encode messages as environmental stimuli influencing the motions of the robotic agent and to decode messages from the resulting motion trajectory. The constraints of maximal robot integrity and minimal motion deviation are established as fundamental principles underlying secrecy. As a proof of concept, we conduct experiments in simulated environments across various manipulation tasks, incorporating robotic embodiments equipped with generalist multimodal policies. 
\end{abstract}


\section{Introduction}
\IEEEPARstart{S}{teganography}, the art of concealing information within non-suspicious media, is rooted in the exchange of messages, which has always carried with it the timeless challenge of secrecy~\cite{10.1007/3-540-61996-8_27, 4655281, 668971, 771065, 935180, Fridrich:2009aa}. From whispers in shadows to the hidden messages written in the margins of history, humankind has long sought ways to convey thoughts that remain imperceptible to all but the chosen few. This ancient pursuit of covert communication has evolved, stretching across the realms of visual, auditory, and linguistic media~\cite{10.1080/0161-119291866883, 10.1007/3-540-61996-8_48, 5387237, 1511007, chang-clark-2014-practical, Zhu:2018aa, ziegler-etal-2019-neural}. In the intricate patterns of imagery, the subtle modulation of sound and the carefully crafted structures of language, steganography advances in various forms, continually adapting to the evolving steganalytic detection mechanisms that seeks to reveal hidden messages~\cite{10.1007/10719724_5, 1040098, 1203220, 10.1145/1411328.1411349, 6081929, 6197267, 7444146}.

This study embarks on an exploration of a new steganographic paradigm in robotics, the realm where the cyber and physical worlds intersect, expanding the boundaries of what is considered a viable channel for covert communication. We consider a form of steganography through the very motions of a robotic agent. Robotics is an interdisciplinary study dedicated to the pursuit of intelligent behaviours that mimic human actions~\cite{10.5555/96732}. At the heart of this quest lies robotic motion control, a domain that spans automation from basic manipulation to complex interaction with dynamic environments~\cite{10.1145/174147.174150}. The integration of artificial intelligence, particularly reinforcement learning, has endowed robotic agents with a high level of autonomy, enabling them to learn, adapt and optimise their decision-making policies over time~\cite{sutton1998reinforcement, Mnih:2015aa, Collins:2024aa}.

It is conceivable to design learning algorithms that adjust a robot’s policies for the purpose of steganography, guiding it to embed hidden messages within its very motions. However, a legitimate robot may have integrity regulations in place to protect the fundamental ethical and safety guidelines programmed into its underlying control model~\cite{Floridi:2018aa, 10.1145/3306618.3314289, 8662743}. Any unauthorised attempts to modify the robot might not go unnoticed. Such illicit manipulations could trigger alarms and activate built-in safeguard mechanisms to prevent catastrophic consequences. Thus, we propose a research challenge, one that requires steganographic methodology in robotics to adhere to the constraint of perfect robot integrity.

In this study, we introduce a steganographic methodology in the context of robotic motion control, subject to the constraint of perfect robot integrity. We exploit the robot’s sensitivity to environmental fluctuations, representing messages as subtle deviations in its motion trajectory. The causal relationships between the influencing factors and the message symbols are established through a trial-and-error heuristic in a simulated virtual environment, until each message symbol is uniquely represented. The synchronised encoding and decoding processes are then applied in the physical world, where the robot’s motion trajectory serves as the medium for transmitting secret information.

The remainder of this paper is organised as follows. Section~\ref{sec:concept} outlines the key concepts in robotic motion control that underpin this study. Section~\ref{sec:method} formalises the problem and methodology for steganography in robotic motion control. Section~\ref{sec:theory} explores the statistical aspects of capacity settings through theoretical analysis. Section~\ref{sec:experiment} presents the experimental validation across various motion control tasks and multimodal robotic agents, including visualisations of simulated environments and performance evaluations on capacity, secrecy and efficiency. Finally, Section~\ref{sec:conclusion} concludes the paper with a summary of research findings and potential future directions.

\section{Foundations of Robotics}\label{sec:concept}
This section provides a brief introduction to fundamental concepts and terminology in robotic motion control, which are relevant to the development of the steganographic methodology.

\subsection{Dynamical Simulation}
The field of robotics has been propelled by the quest for developing \emph{generalist robotic policies} capable of handling a broad spectrum of tasks, including manipulation, grasping and assembly, with precision and autonomy~\cite{10.5555/561828}. To evaluate robotic policies, \emph{real-world benchmarks} are crucial as they provide assessments in uncontrolled environments, with varying environmental factors such as lighting conditions, material properties and sensor errors~\cite{7583659}. However, real-world evaluation is hindered by several limitations, including resource-demanding and time-consuming setup and maintenance. Additionally, it faces challenges in scalability and reproducibility, as testing a wide variety of conditions efficiently and ensuring consistent experimental conditions in dynamic real-world environments can be difficult. 

These constraints become particularly evident as robotic systems are deployed in complex environments, driving a shift toward \emph{simulation-based evaluations} as a more cost-efficient, scalable and reproducible alternative. Simulators equipped with \emph{physics engines} offer a sufficiently realistic approximation of physical systems, capturing the essential dynamics that allow robots to be evaluated under controlled virtual conditions~\cite{1389727, 6386109, 6696520, makoviychuk2021isaac}. These simulations can mimic the dynamics of real-world interactions, enabling the evaluation of robotic policies across various scenarios with unlimited trails. The flexibility to subtly manipulate various environmental factors, coupled with the capability for repeated trials, provides an ideal platform for the proposed steganographic methodology.

\subsection{Mechanical Robot}
A robot is a physical agent equipped with sensors, actuators and computational resources. It interacts with the environment by perceiving states through its sensors and taking actions autonomously via its actuators~\cite{1014739}. A robot's actions are governed by an underlying policy model, which maps the sensed state to an optimal action~\cite{BELLMAN:1957aa}. In reinforcement learning, this policy is often learned through interactions with the environment to maximise cumulative rewards~\cite{Sutton:1988aa, watkins1989learning, Lin:1992aa, Williams:1992aa, NIPS1999_464d828b}. The development of multimodal robotic agents has advanced robotic motion control through the integration of multiple sensory modalities such as auditory, visual and linguistic information~\cite{10.5555/3104482.3104569, JMLR:v15:srivastava14b, 8269806, 10.1145/3656580}. These multimodal sensors enhance contextual awareness and human-machine interaction, enabling robots to respond more intelligently and interact more naturally with human operators. For example, a visual-language-action robotic agent can process visual scenes to understand the context of the environment and interpret linguistic commands from humans to execute specific actions

A robotic arm is a mechanical system designed to replicate the functions of a human arm, serving as an ideal example for demonstrating robotic motion control~\cite{1087068}. These arms are widely used in industrial applications due to their versatility in executing complex manoeuvres. A common type of robotic arm, designed with 7 degrees of freedom, exhibits motions that can be generally categorised into the following primary types:
\begin{itemize}
	\item Positional Motion: The motion that moves the position of the end effector in a linear direction along predefined axes (X, Y and Z).
	\item Rotational Motion: The motion that adjusts the orientation of the end effector (pitch, yaw and roll).
	\item Functional Motion: The motion that controls the gripping mechanism of the end effector such as claws or fingers to grasp or release.
\end{itemize}
This study aims to develop a steganographic methodology capable of seamlessly operating within scenarios involving modern robotic agents.

\begin{figure*}[!t]
\begin{center}
\includegraphics[width=1.6\columnwidth]{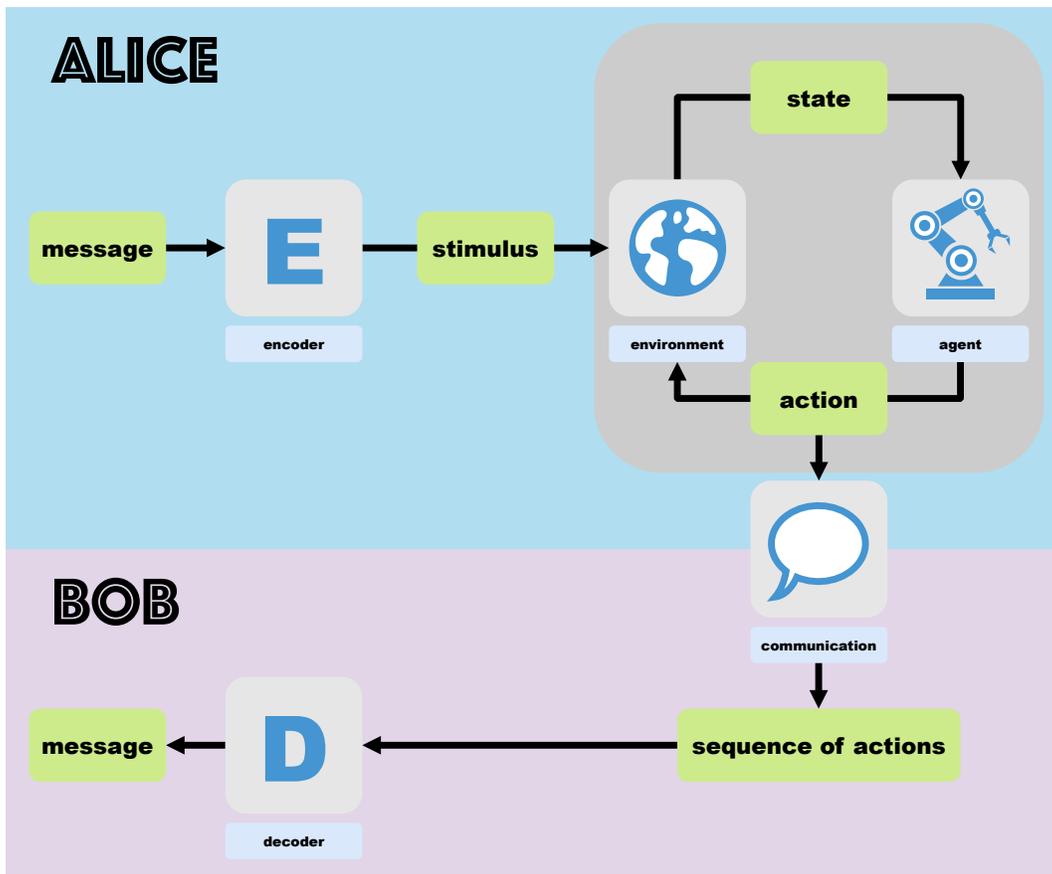}
\end{center}
\caption{Overview of steganography in robotic motion control: Alice encodes a message as a stimulus that affects the interaction between the robotic agent and the environment, whereas Bob decodes the message from the sequence of actions transmitted via the robot's in-built communication system.}
\label{fig:overview}
\end{figure*}

\section{Steganography in Robotics}\label{sec:method}
This section outlines the methodology for steganography in the context of robotic motion control. We begin by presenting the problem formulation, which defines the core components and constraints of the research. Following that, a trial-and-error heuristic is introduced, leveraging the robot's sensitivity to environmental changes for covert communication of secret information

\subsection{Problem Formulation}

We present a formal problem of steganography applied to robotics, where the objective is to covertly encode a secret message into the motions of a robot. The sender Alice influences the robot's motions while it is tasked with performing a predefined mission. The receiver Bob observes the motion signal from a remote location and decodes the secret message embedded in the motions. This subliminal communication should not significantly interfere with the robot’s normal operations for the secrecy requirement.
    
Formally, let the robot's motion trajectory be defined as a sequence of actions $ \mathbf{a} = \{ a_1, a_2, \dots, a_T \} $, where $ a_t \in \mathbb{R}^n $ represents the configuration of the arm at time-step $ t $, with $ T $ being the total number of time steps. For instance, the action at each time-step can correspond to the 7 degrees of freedom, assuming a 7-axis robotic arm. The robot’s decision process is governed by a pre-trained control model, which determines how the robot acts and moves from an initial state to a terminal state. Alice, who is located in the same environment as the robot, can stimulate subtle deviations in the motion trajectory to encode an intended secret message $ m $. Bob, located remotely, can observe the motion trajectory $ \mathbf{a} $ from which the message $ m $ is decoded. The motion signal is transmitted as part of the robot's built-in communication system, which is designed for real-time monitoring and tracking of the robot's operations to support diagnostics and ensure safety. Some sensitive data, such as camera scenes, may be unavailable for analysis due to privacy regulations. In this context, Alice exploits the robot's surveillance system to communicate secretly with Bob, who disguises himself as an authorised observer and gains access to the motion signal under the pretence of performing legitimate maintenance duties. In this steganographic framework, the following core constraints are applied.
\begin{itemize}
    \item Maximal Robot Integrity: The robot’s underlying control model is protected from unauthorised tampering or retraining, which could lead to malfunction or unintended behaviours, thereby jeopardising the safety of the environment in which the robot operates. Any attempt to modify the predefined model parameters may be detected, triggering alarms or other built-in safeguard mechanisms.
    
    \item Minimal Motion Deviation: The robot’s motion trajectory is required to align with the expected motion statistics. An excessive deviation from the typical trajectory may signal a malfunction or raise steganalytic suspicions, triggering an automatic halt in operation for further investigation.
\end{itemize}
Note that while cyber-physical gaps and observational errors, which may arise due to simulation imperfection, actuation precision, communication noise or sensor limitation, are not always negligible, the scope of this study is focused on a prototypical setting to establish foundational principles.

\subsection{Trail-and-Error Heuristic}
The robot's control model can be sensitive to \emph{environmental stimuli}, meaning that changes in the surroundings, such as the introduction of new objects or variations in the background, can affect its decision-making processes. These changes may lead to deviations in the robot's planned trajectory or force it to adapt its actions to accommodate and account for the new environmental factors. On the one hand, such sensitivity could potentially enhance the robot's ability to operate in dynamic environments; on the other hand, it also exposes potential vulnerabilities, as unforeseen environmental changes can cause unintended consequences that disrupt the robot's normal functioning.

In view of this, we propose a trial-and-error heuristic that exploits the robot's sensitivity to environmental changes for encoding and decoding hidden messages, as illustrated in Figure~\ref{fig:overview}. Initially, Alice and Bob agree upon a common \textit{stego-key}, which serves as the seed for initialising a shared decoder, ensuring the synchronisation of encoding and decoding processes. According to Kerckhoffs's principle and Shannon's maxim in cryptography, one ought to assume that the enemy knows the system~\cite{6769090}. A system should be secure, even if everything about the system, except the key, is public knowledge. In other words, the security of a system should not rely on the secrecy of the algorithm, but rather on the secrecy of the key. In a steganographic system, the security is governed by a stego-key, which coordinates the operations between the two parties, Alice and Bob.

Alice exploits the robot's sensitivity to environmental changes by sampling and placing subtle stimuli in its environment, which influence the robot’s motion trajectory. Each resulting motion trajectory can be mapped to a message symbol by the shared decoder, which, being a many-to-one function, may produce duplicate symbols. Therefore, the process is iterated (in a simulated cyber environment), following a trial-and-error heuristic, with stimuli being resampled until all unique message symbols are represented. Once the set of stimuli corresponding to the entire symbol space has been established, these stimuli are applied (in the physical world) to encode any secret message, which is then communicated in the form of a motion trajectory. Bob, located remotely, observes the motion trajectory and uses the shared decoder to extract the hidden message. This method preserves the integrity of the robot's control model while allowing for asymptotically minimal deviation as the number of trials increases. A step-by-step methodology is outlined as follows.

\paragraph*{\textbullet\ Initialisation of Decoder: Alice and Bob exchange a common stego-key, from which the shared decoder is initialised, as denoted by
\begin{equation}
	\mathcal{D} = \operatorname{init}(k) .
\end{equation}
This decoder function serves as a mapping from the action sequence to the message space $\mathcal{M}$.
}

\paragraph*{\textbullet\ Initialisation of Encoder: Alice randomly samples a stimulus $\psi$ and places it within the simulated cyber environment, causing a slight environmental change. The sequence of actions is generated though the interactions between the robot and its environment in the presence of the stimulus $\psi$, as given by
\begin{equation}
	\mathbf{a} = \operatorname{interact}(\psi) .
\end{equation}
This action sequence is then mapped to a message symbol by the shared decoder. The process is iterative, with new stimuli being resampled and trials executed until the set of decoded symbols covers all possible symbols in the message space $\mathcal{M}$. Therefore, each distinct message symbol can be encoded as an environmental stimulus, as represented by
\begin{equation}
	\psi_{m} = \mathcal{E}(m) .
\end{equation}
For symbols that correspond to more than one stimulus, the stimulus resulting in the optimal motion trajectory (which minimises time costs or maximises efficiency) can be chosen.
}

\paragraph*{\textbullet\ Encoding process: On the transmitting side, Alice encodes the intended message as its corresponding stimulus and positions it in the physical environment, influencing the robot's motion during its interactions with the environment, as expressed by
\begin{equation}
	\mathbf{a} = \operatorname{interact}(\mathcal{E}(m)) .
\end{equation}
}

\paragraph*{\textbullet\ Decoding process: At the receiving end, Bob applies the shared decoder to the observed motion trajectory and extracts the hidden message by
\begin{equation}
	m = \mathcal{D}(\mathbf{a}) .
\end{equation}
}

\begin{table}[t!]
\centering
\caption{Comparison of empirical, theoretical and approximate means.}
\label{tab:mean_comparison}
% \begin{tabular}{@{}c | ccc@{}}
\begin{tabular}{c | c  c  c}
\toprule
\toprule
\textbf{\( n \)} & \textbf{Empirical Mean} & \textbf{Theoretical Mean} & \textbf{Approximate Mean} \\ 
\midrule
2  & 02.99 & 03.00 & 02.54 \\
\midrule
3  & 05.51 & 05.50 & 05.03 \\
\midrule
4  & 08.37 & 08.33 & 07.85 \\
\midrule
5  & 11.32 & 11.42 & 10.93 \\
\midrule
6  & 14.57 & 14.70 & 14.21 \\
\midrule
7  & 18.22 & 18.15 & 17.66 \\ 
\midrule
8  & 21.67 & 21.74 & 21.25 \\ 
\bottomrule
\bottomrule
\end{tabular}
\end{table}

\section{Theoretical Analysis}\label{sec:theory}
The code construction begins by fixing the decoder, and then establishes the encoder by randomly sampling codes (stimuli) until every symbol in the message space is uniquely represented by at least one sampled code. The capacity is calculated as the binary logarithm of the message space size, $\log_2 \| \mathcal{M} \|$, with the unit expressed in bits per trajectory. The main challenge in this process lies in determining how many random codes must be sampled to achieve complete coverage of all message symbols with high confidence. This section seeks to answer questions of how many trials are needed to ensure success with high confidence and how likely is complete coverage after a given number of trials.

\subsection{Random Coding}
The trial-and-error heuristic can be implemented using a decoder constructed with hashing and modular arithmetic. Specifically, the decoder employs a keyed hash function followed by modular reduction to ensure that all possible codes are confined to the fixed range of the message space, as expressed by
\begin{equation}
	m = \operatorname{hash}(\psi_{m}) \mod \|\mathcal{M}\|
\end{equation}
 This decoder deterministically maps each sampled code to a symbol within the message space. The uniformity of the hash function minimises the likelihood of uneven distribution, ensuring that symbols in the message space are approximately equally represented. This uniformity allows for straightforward theoretical analysis of the relationships between the number of trials and the full coverage the message space. Note that, despite its theoretical simplicity, this decoding approach is sensitive to errors which might arise in practical implementations. Although addressing the robustness limitations lies beyond the scope of this study,  further investigation into a more adaptable decoding mechanism could enhance resilience against errors.

 \begin{figure}[t]
\begin{center}
\includegraphics[width=1.0\columnwidth]{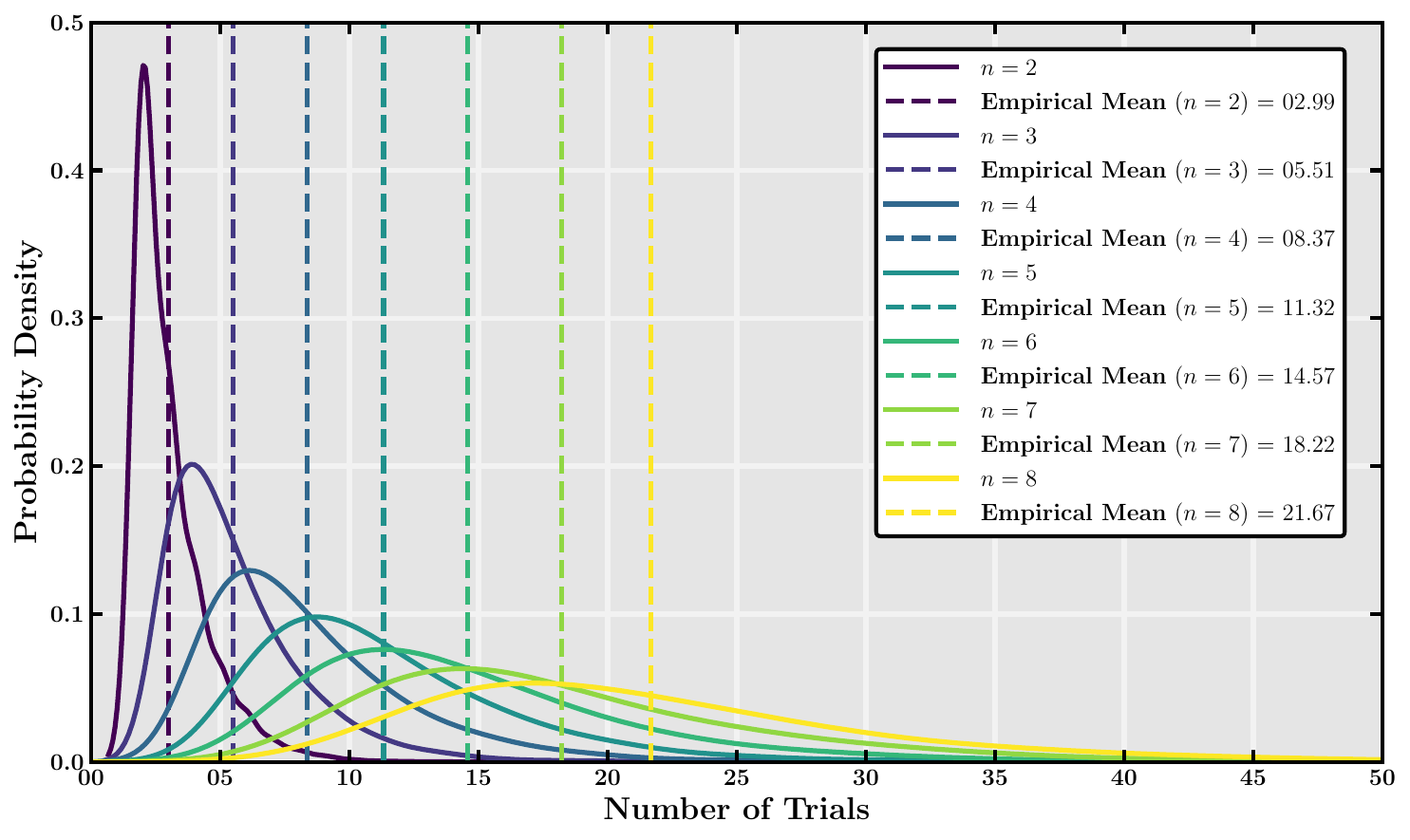}
\end{center}
\caption{Expected number of trials required for complete coverage of message space.}
\label{fig:theory_mean}
\end{figure}

\subsection{Expected Number of Trials}
The problem of determining the required number of random codes to cover all $n$ message symbols is analogous to the \emph{coupon collector’s problem} in probability theory. This problem concerns the expected time (number of trials) to collect all $n$ distinct items (coupons), assuming each trial independently and uniformly selects one item. This expected value is given by
\begin{equation}
	\mathbb{E}[T_n] =  n \cdot \left( 1 + \frac{1}{2} + \frac{1}{3} + \cdots + \frac{1}{n} \right) = n \cdot H_n ,
\end{equation}
where where $H_n$ is the $n$-th harmonic number (the sum of the reciprocals of the first $n$ natural numbers). The $n$-th harmonic number is about as large as the natural logarithm of $n$ by approximating the sum with the integral:
\begin{equation}
	H_n = \sum_{k=1}^n \frac{1}{k} \approx \int_{1}^n \frac{1}{x} dx = \ln n,
\end{equation}
As $n$ goes to infinity, the difference between the harmonic series and the natural logarithm approaches asymptotically towards the limit known as the Euler-Mascheroni constant:
\begin{equation}
	\lim_{n \to \infty} \left( H_{n} - \ln n \right) = \gamma .
\end{equation}
Given the asymptotic behaviour of the harmonic number, the expected value can be approximated as
\begin{equation}
	n \cdot H_{n} \sim n \cdot \left( \ln n + \gamma \right).
\end{equation}
This results characterises the expected behaviour of the sampling process and highlights that collecting the final few items takes disproportionately longer than the initial ones, a phenomenon often referred to as the \emph{diminishing returns}. Table~\ref{tab:mean_comparison} compares the mean values across different message space sizes, obtained from empirical simulation, theoretical derivation and asymptotic approximation. Figure~\ref{fig:theory_mean} illustrates the empirical mean of the required trials computed over 10,000 simulations, with the distribution approximated using kernel density estimation on the empirical data.

\begin{figure}[t]
\begin{center}
\includegraphics[width=1.0\columnwidth]{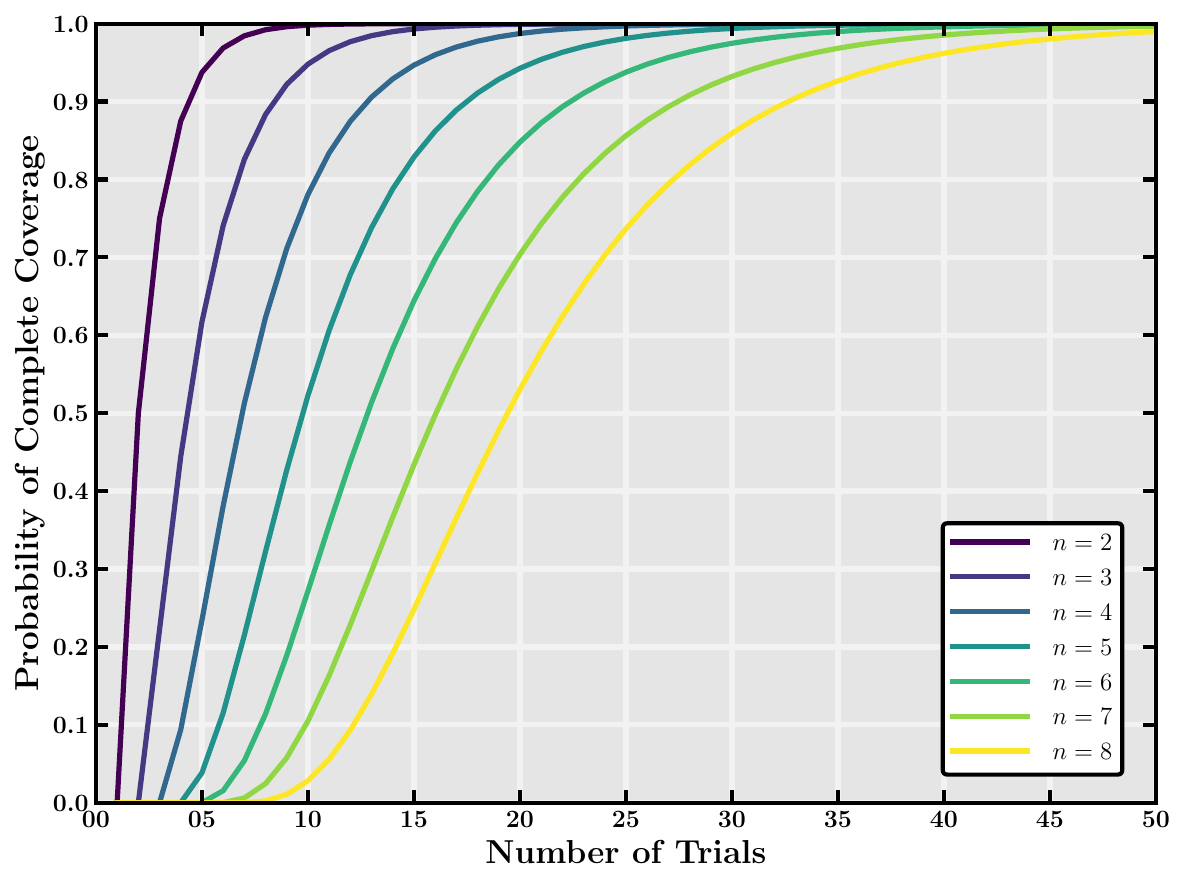}
\end{center}
\caption{Probability of complete coverage with respect to the number of trials conducted.}
\label{fig:theory_prob}
\end{figure}

\begin{figure*}[t!]
    \centering
    % \captionsetup[subfigure]{labelformat=empty}
    % First subfigure
    \subfloat[close middle drawer.]{
    	\includegraphics[width=0.94\columnwidth]{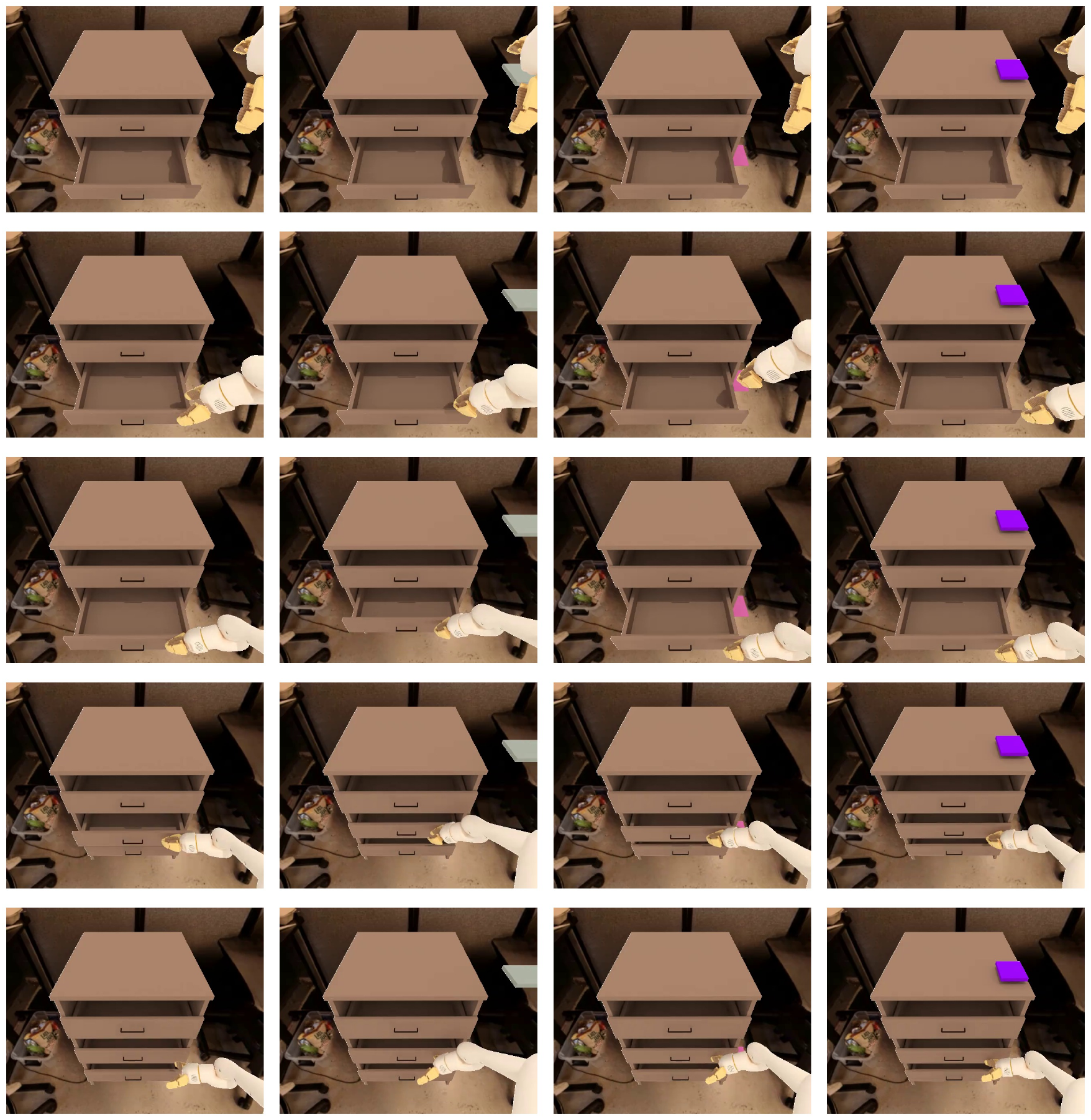}
    }
    \hfil
    % Second subfigure
    \subfloat[move redbull can near apple.]{
    	\includegraphics[width=0.94\columnwidth]{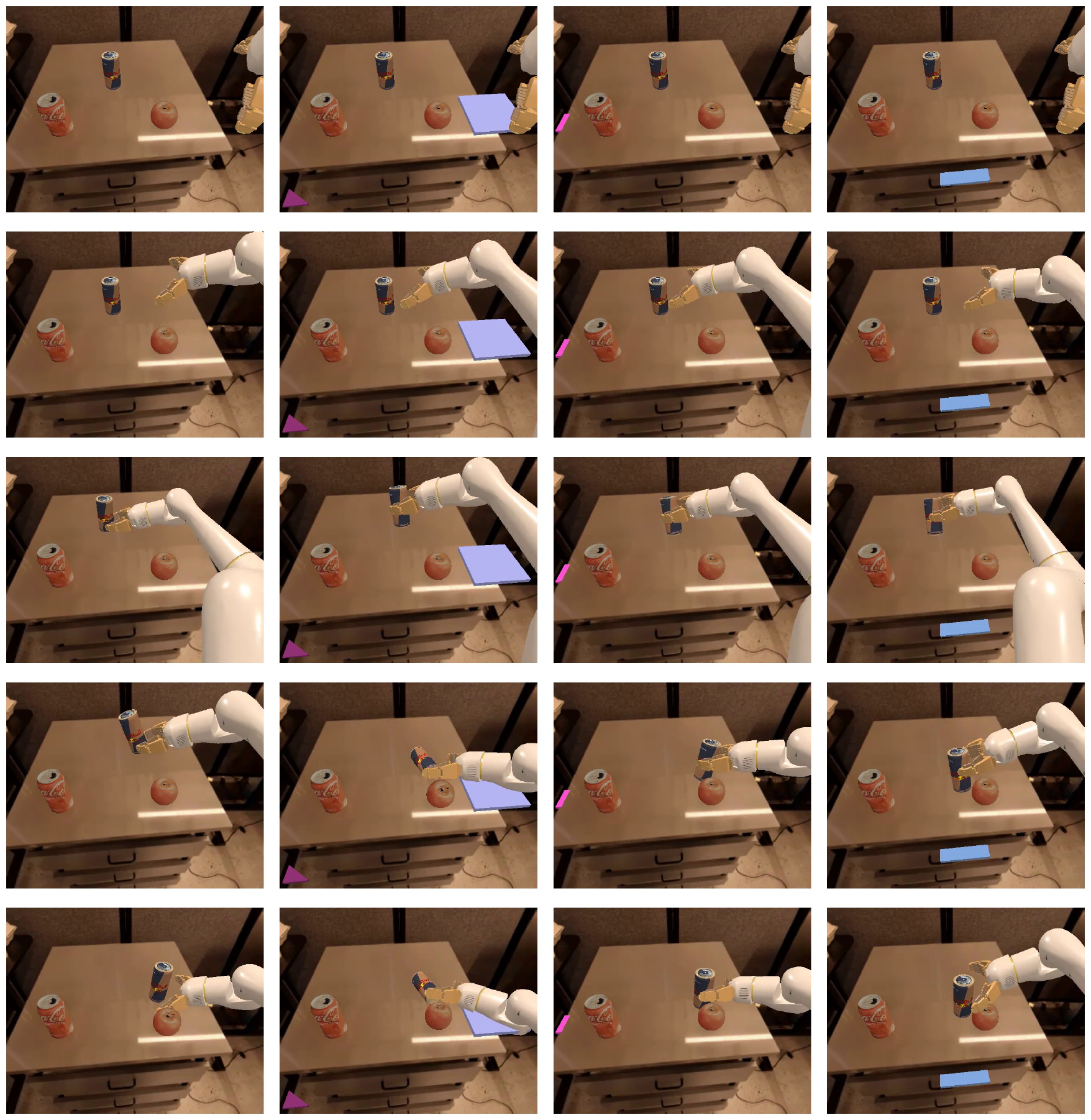}
    }
    \\
    % Third subfigure
    \subfloat[pick coke can.]{
    	\includegraphics[width=0.94\columnwidth]{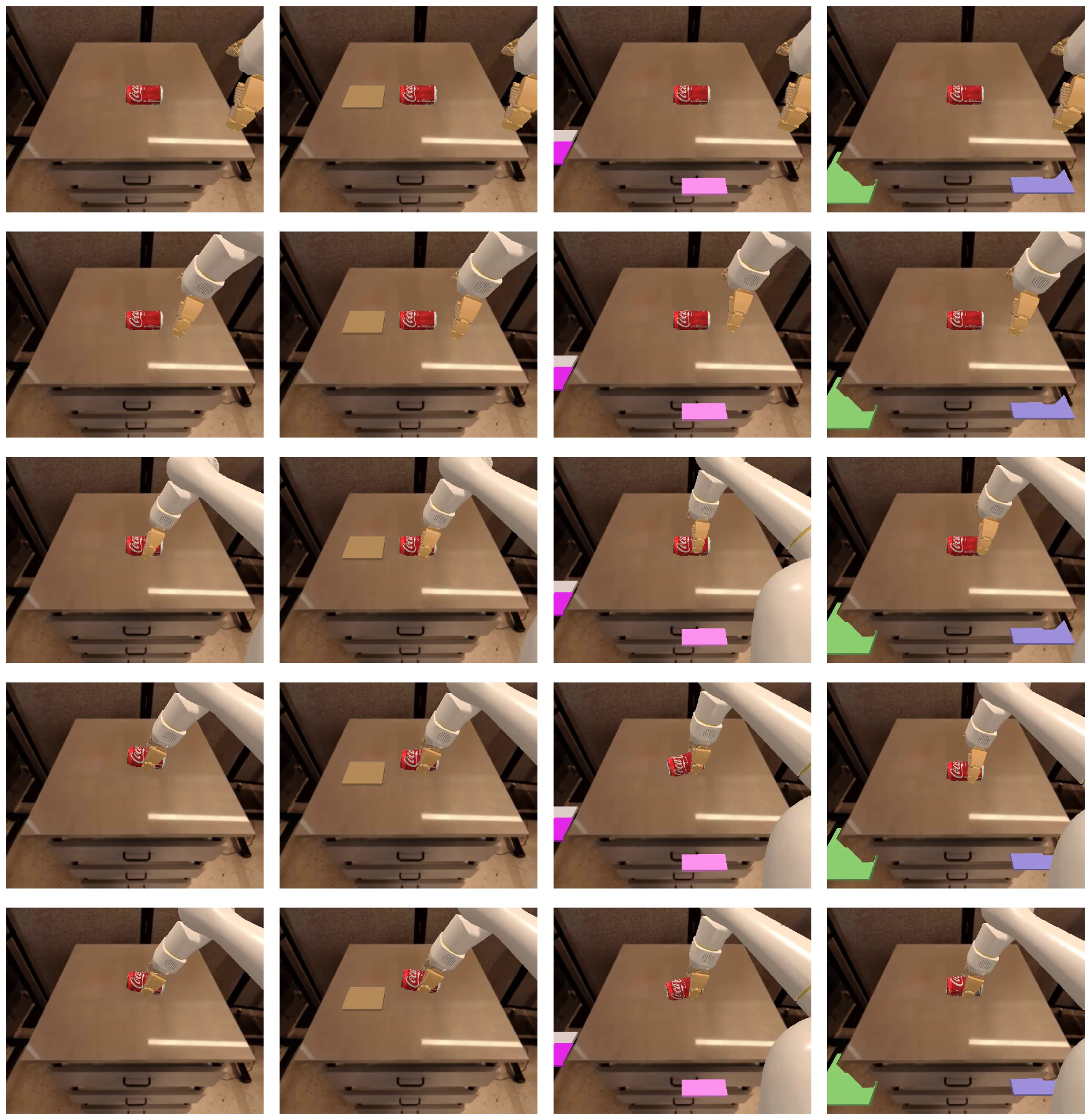}
    }
    \hfil
    % Fourth subfigure
    \subfloat[put aubergine into yellow basket.]{
    	\includegraphics[width=0.94\columnwidth]{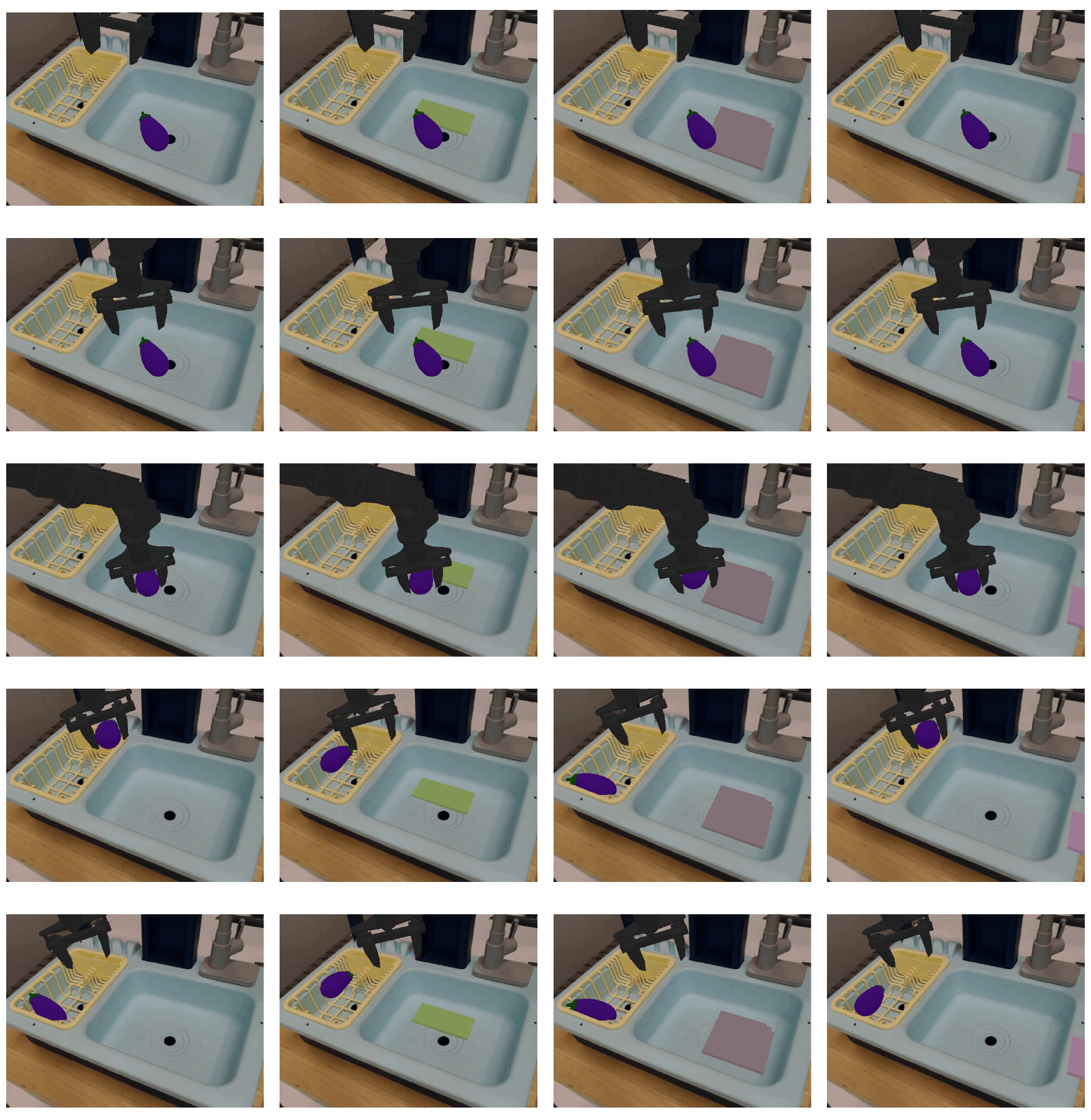}
    }
    \caption{Visualisation of robotic motion control: each column represents a motion trajectory with the non-stimulated one shown in the first column and steganographic variations in the rest, whereas each row represents a sequential time-step.}
    \label{fig:visualisation}
\end{figure*}

\subsection{Probability of Complete Coverage}
Another central question concerns the probability of complete coverage after $t$ trials. Instead of directly calculating the probability that all $n$ items are drawn, we begin by finding the probability that at least one item is not drawn after $t$ trials and then subtract this from 1. For at least one item to remain uncollected, there must exist a subset of $k$ items (where $1 \leq k \leq n$) that are not collected at all. The probability that exactly $k$ specific items are not drawn after $t$ trials is given by
\begin{equation}
	\binom{n}{k} \left( \frac{n-k}{n} \right)^t .
\end{equation}
When calculating the probability that at least one item is not drawn, overlaps between subsets of uncollected items must be carefully accounted for. The \emph{inclusion-exclusion principle} addresses this by alternately adding and subtracting probabilities of subsets of increasing size. Starting with the probability of missing single items, we subtract probabilities for pairs of missing items to correct overcounting, then add probabilities for triples to correct undercounting, and so on. This alternating process handles all overlaps, leading to the probability that at least one item is not drawn after $t$ trials. Finally, subtracting this from 1 gives the likelihood of achieving complete coverage in a finite number of trials:
\begin{equation}
	\mathbb{P}(t) = 1 - \sum_{k=1}^{n} (-1)^{k+1} \binom{n}{k} \left( \frac{n-k}{n} \right)^t .
\end{equation}
Figure~\ref{fig:theory_prob} illustrates the probability of achieving complete coverage as a function of the number of trials, demonstrating how a larger number of items requires more trials to reach a comparable probability.

\begin{figure*}[t!]
    \centering
    % \captionsetup[subfigure]{labelformat=empty}
    % First subfigure
    \subfloat[close middle drawer.]{
    	\includegraphics[width=0.98\columnwidth]{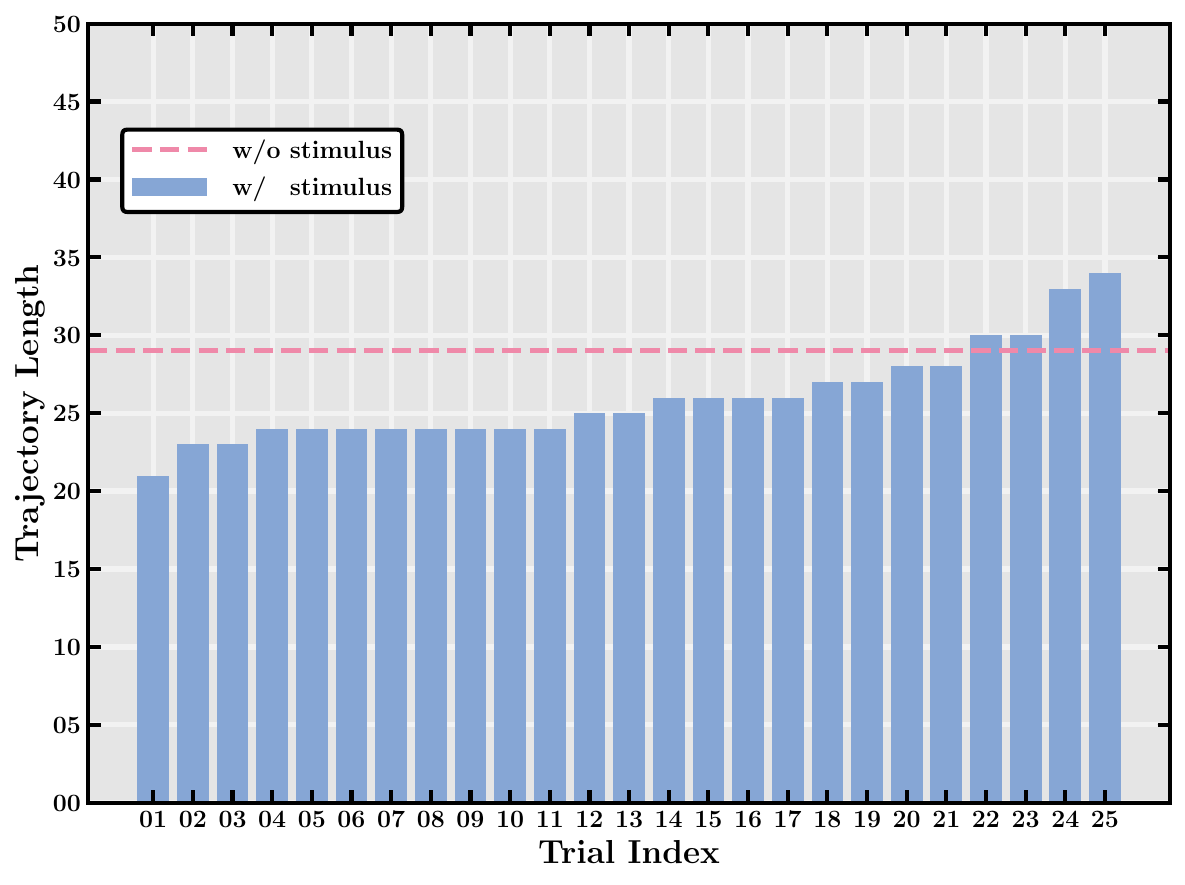}
    }
    \hfil
    % Second subfigure
    \subfloat[move redbull can near apple.]{
        \includegraphics[width=0.98\columnwidth]{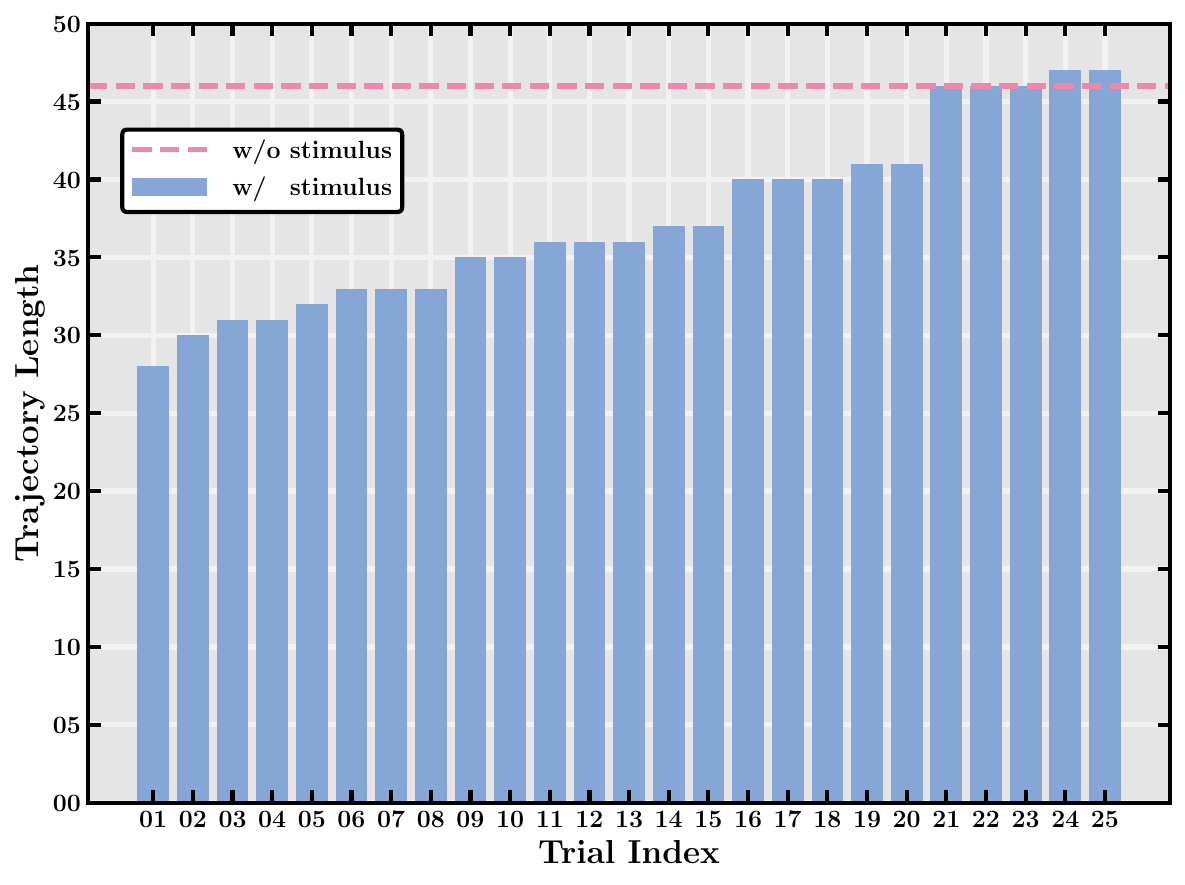}
    }
    \\
    % Third subfigure
    \subfloat[pick coke can.]{
        \includegraphics[width=0.98\columnwidth]{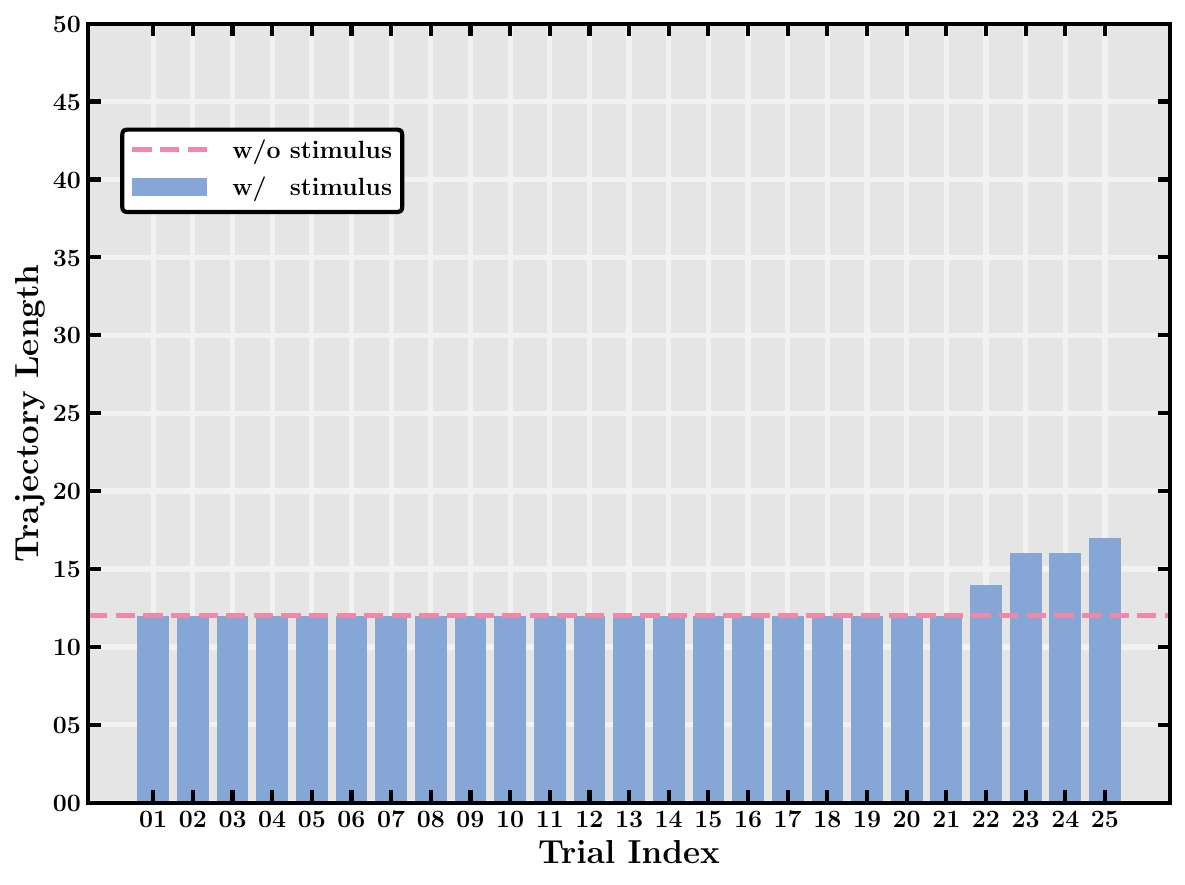}
    }
    \hfil
    % Fourth subfigure
    \subfloat[put aubergine into yellow basket.]{
        \includegraphics[width=0.98\columnwidth]{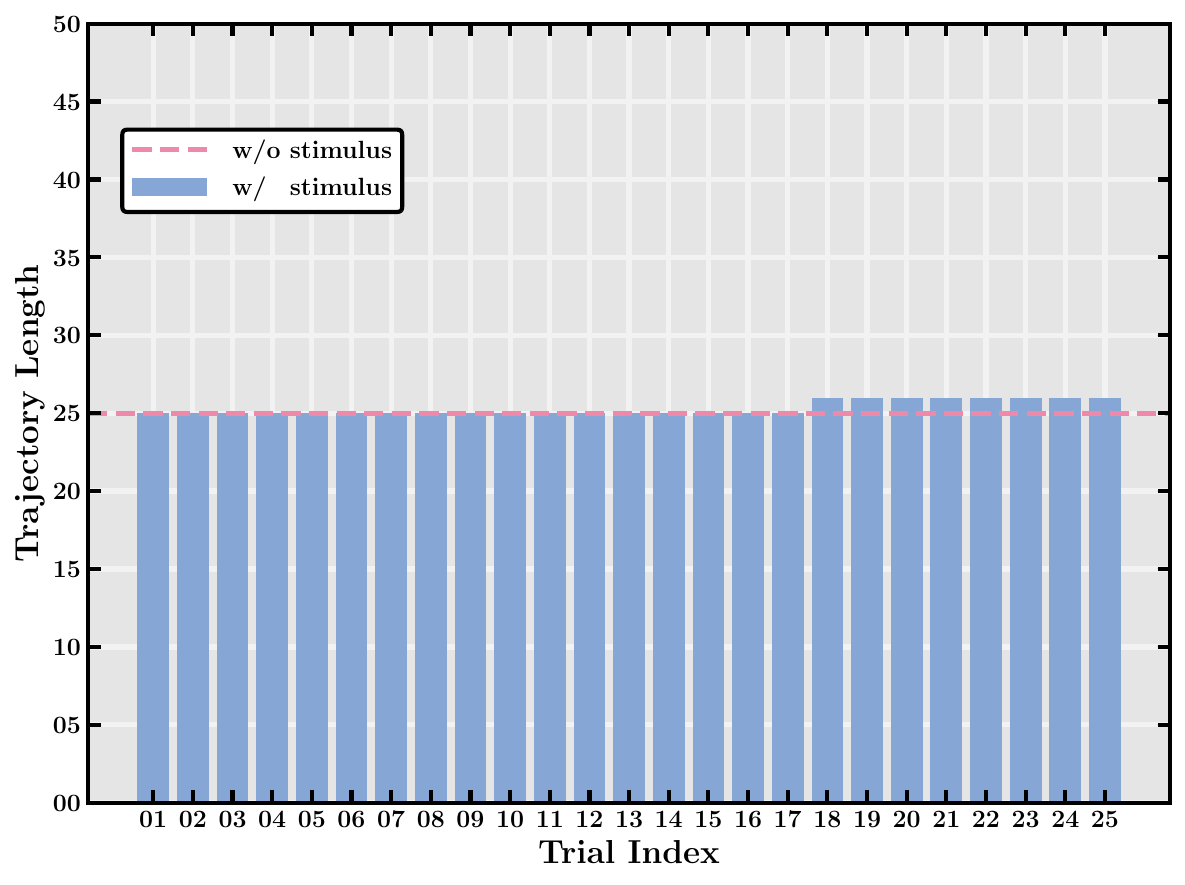}
    }
    \caption{Secrecy analysis based on the statistics of trajectory lengths.}
    \label{fig:secrecy}
\end{figure*}

\begin{figure*}[t!]
    \centering
    % \captionsetup[subfigure]{labelformat=empty}
    % First subfigure
    \subfloat[close middle drawer.]{
    	\includegraphics[width=0.98\columnwidth]{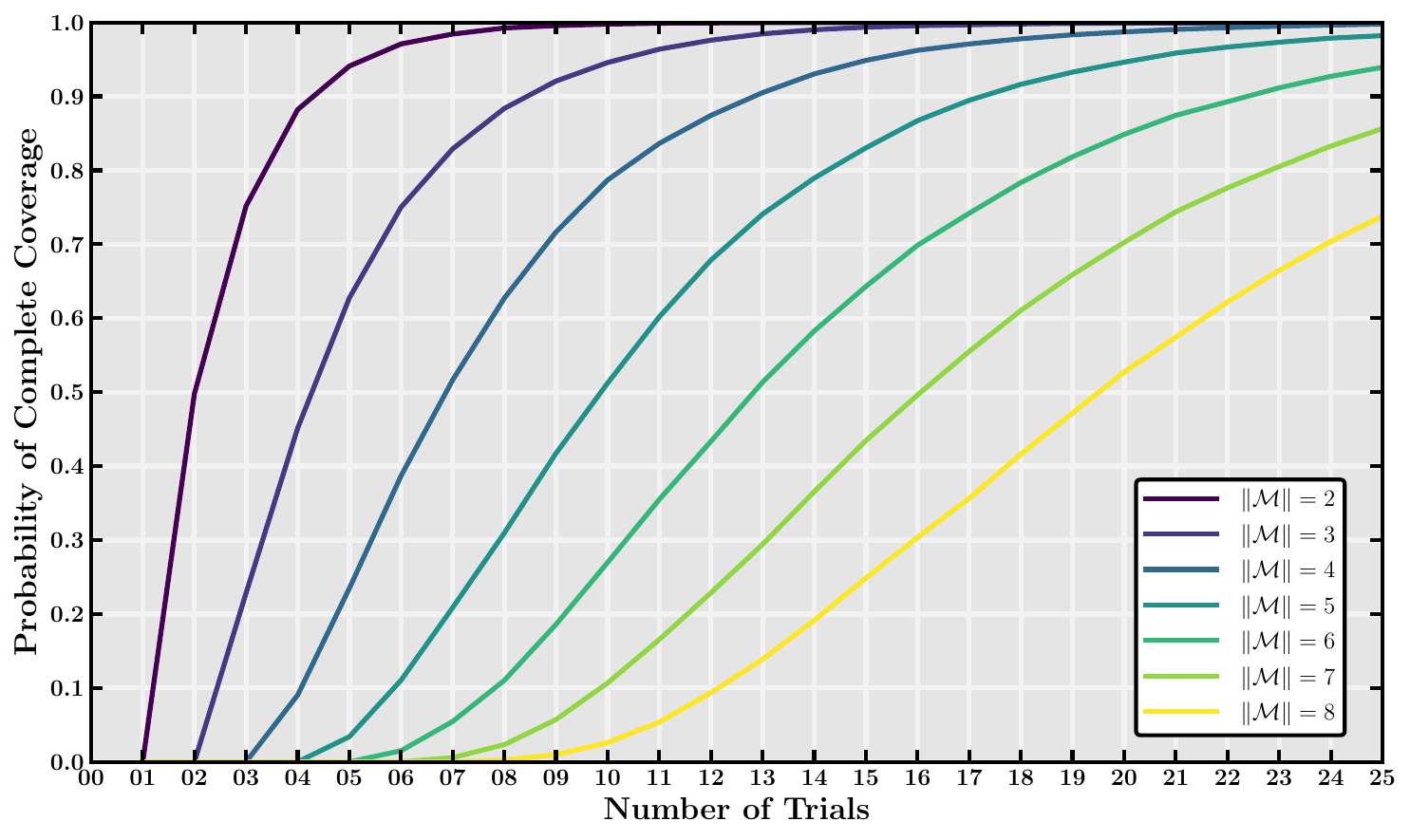}
    }
    \hfil
    % Second subfigure
    \subfloat[move redbull can near apple.]{
        \includegraphics[width=0.98\columnwidth]{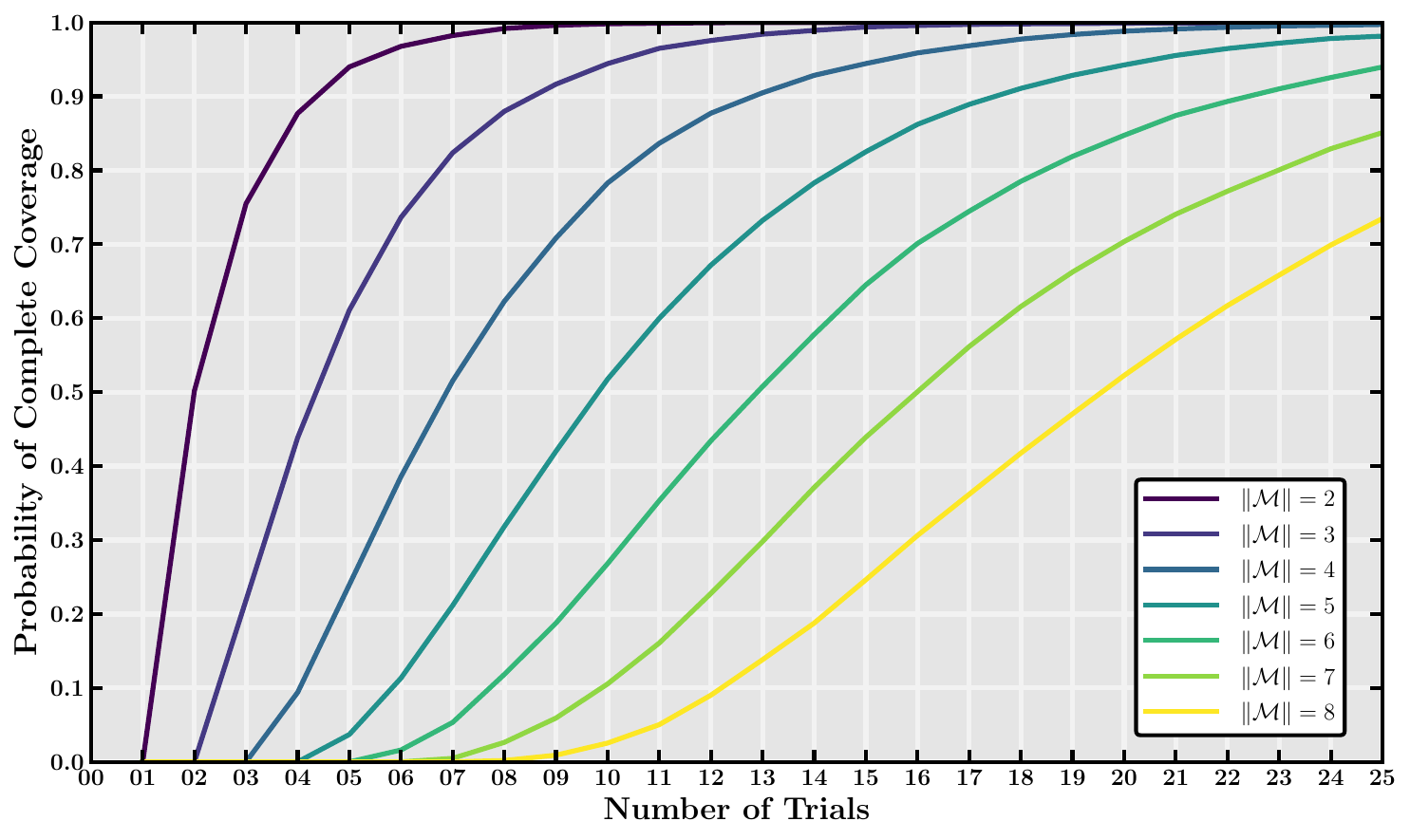}
    }
    \\
    % Third subfigure
    \subfloat[pick coke can.]{
        \includegraphics[width=0.98\columnwidth]{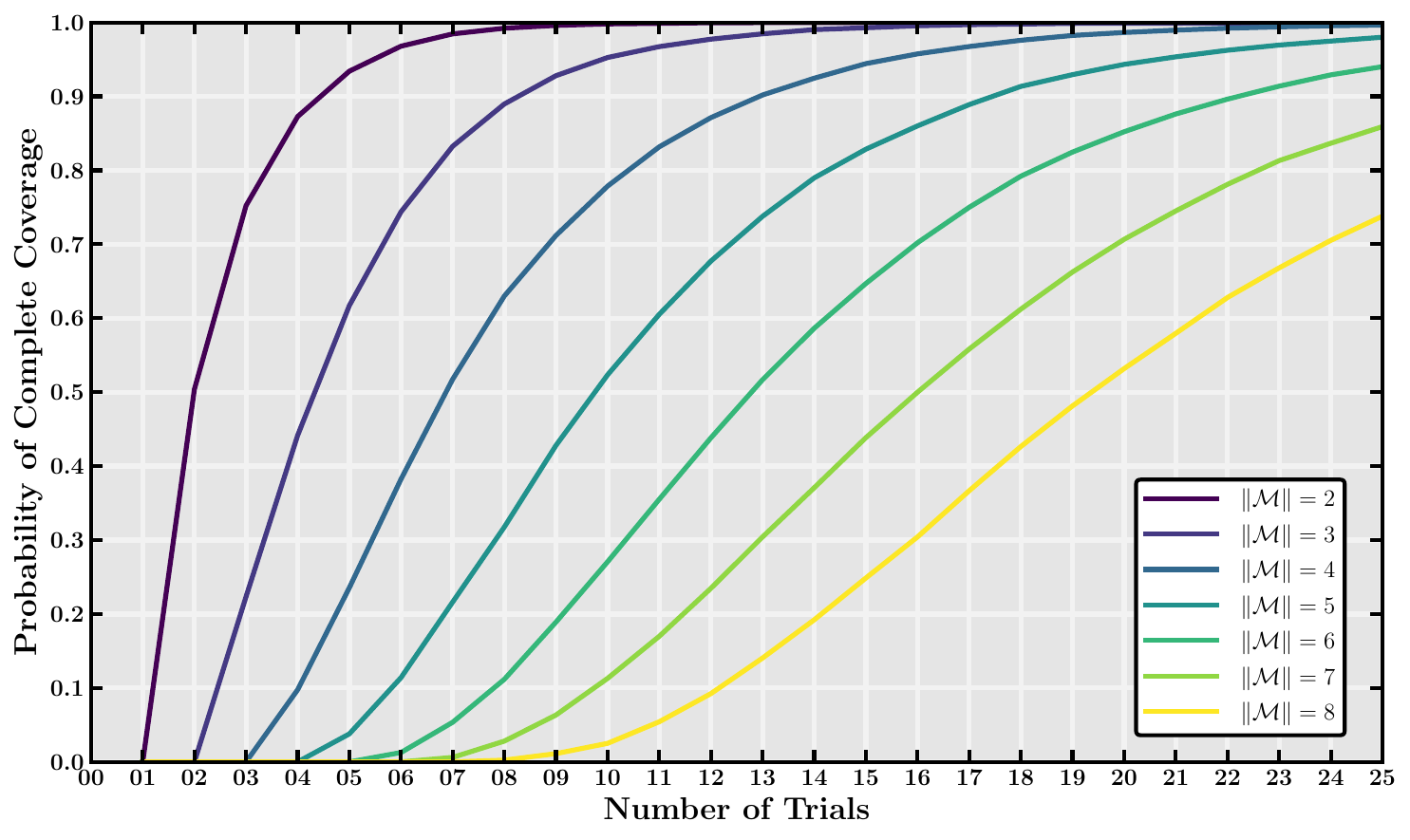}
    }
    \hfil
    % Fourth subfigure
    \subfloat[put aubergine into yellow basket.]{
        \includegraphics[width=0.98\columnwidth]{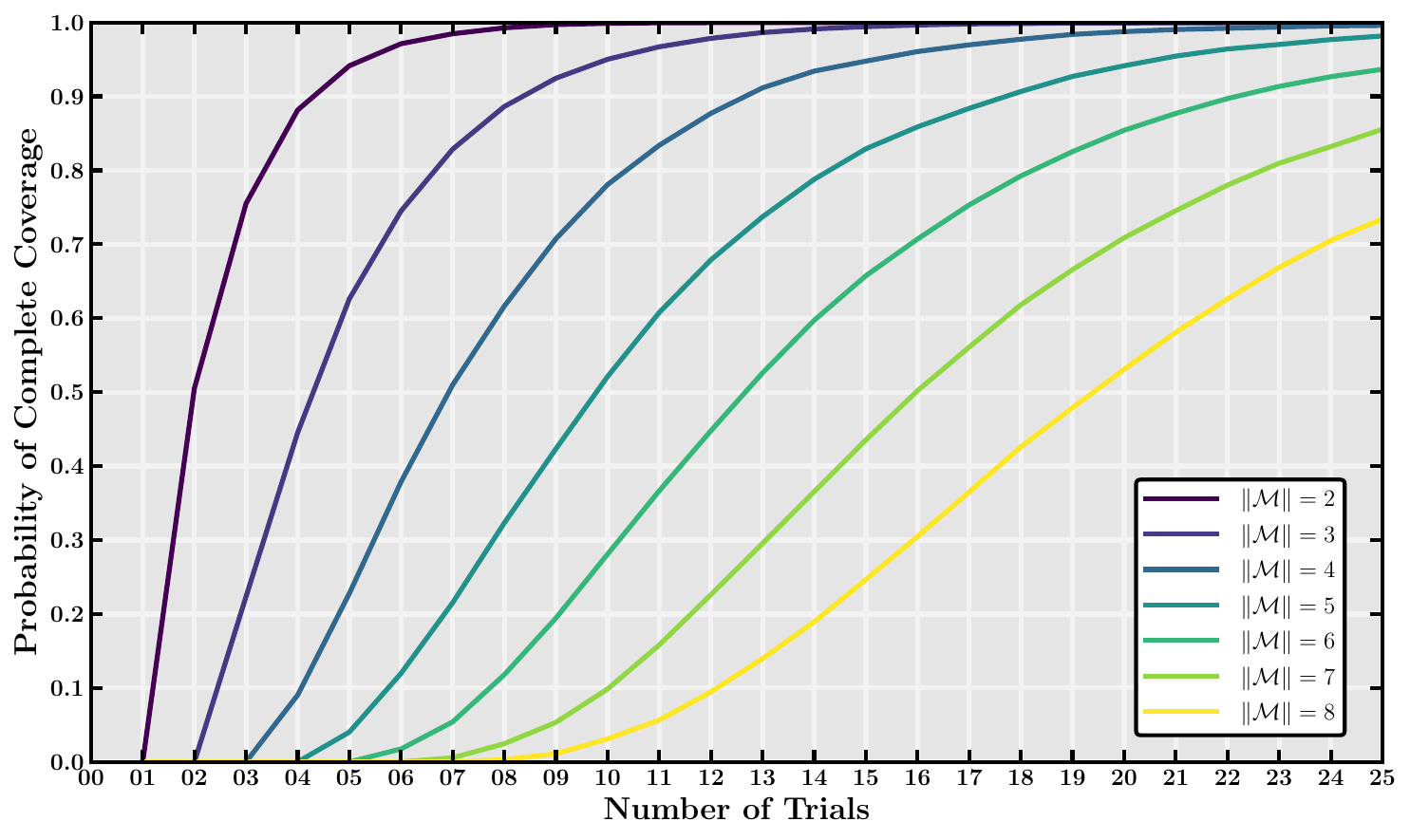}
    }
    \caption{Capacity analysis based on the probability of complete coverage over the message space.}
    \label{fig:capacity}
\end{figure*}

\section{Experiments}\label{sec:experiment}
This section presents experimental evaluations. It begins with a description of the simulation setup, followed by a visualisation of motion trajectories through a simulation interface. The steganographic performance is assessed in terms of both secrecy and capacity.

\subsection{Simulation Setup}
Our evaluations spanned four distinct tasks in a dynamical simulator~\cite{li2024evaluating}: closing the middle drawer, moving a Red Bull can near an apple, picking up a Coke can, and placing an aubergine into a yellow basket. These tasks were performed by two robotic embodiments:  the first three tasks were executed by the Google robotic arm, while the last task was carried out by the WidowX robotic arm. The robotic embodiments were controlled by two multimodal policy models: the motions in the first two tasks were governed by OpenVLA~\cite{kim2024openvla}, while the motions in the latter two tasks were directed by Octo~\cite{Ghosh-RSS-24}. For each task, there were 25 steganographic trajectories generated through trials of sampling random objects as environmental stimuli, in addition to an original trajectory serving as a referential baseline. Following the law of large numbers, the experiments were conducted with 10,000 pseudo-randomly generated keys, each used to initialise a unique decoder.

\subsection{Visualisation}
Figure~\ref{fig:visualisation} visualises the motion trajectories executed by a robotic agent for each task in both conditions, with and without stimuli. This visualisation highlights how environmental stimuli subtly influenced the motion trajectories, with minimal observable deviations between the baseline and steganographic trajectories. While the stimuli were instantiated as monochromatic shape objects placed in random locations to simplify the demonstration of the methodology, we do not impose strict assumptions on the manifestation of stimuli. In practice, stimuli can be arbitrarily and flexibly designed as any objects to suit the contextual secrecy requirements of specific environments.

\subsection{Secrecy Analysis}

We evaluated secrecy via the statistics of trajectory lengths. Figure~\ref{fig:secrecy} depicts the lengths of steganographic trajectories compared to the length of the non-stimulated trajectory. The deviations in trajectory lengths reflect the steganographic impact, which did not result in a monotonic increase or decrease the task completion times. Instead, unpredictable variations were observed in the lengths of steganographic trajectories. While these deviations in trajectory lengths provide an initial indication of the potential inconspicuousness of the steganographic system, this metric alone may not fully capture the comprehensive statistics related to secrecy. To the best of our knowledge, no steganalysis methods have been specifically developed for steganography in robotic motion control at the time of writing, making this an emerging area for further exploration.

\subsection{Capacity Analysis}
We evaluated capacity with respect to the probability of complete coverage over the message space. Figure~\ref{fig:capacity} demonstrates this probability as a function of the number of trials, evaluated across 10,000 randomly initialised decoders. The simulations were conducted for various message space sizes, ranging from 2 to 8 symbols, corresponding to capacity settings of 1 to 3 bits (calculated as binary logarithm of the message space size). The curves highlight the progressive increase in the probability of success with an increasing number of trials. The results underscore the balance between capacity and efficiency, demonstrating that smaller message spaces enable faster convergence, whereas larger spaces necessitate additional trials to maintain high coverage probabilities. 

\section{Conclusions}\label{sec:conclusion}
This study introduced a steganographic paradigm in robotic motion control, exploring the robot's sensitivity to environmental stimuli to allow covert communication via motion trajectories. Following the formulated principles regarding maximal robot integrity and minimal motion deviation, we demonstrated how a message can be encoded as an environmental stimulus that influences the interactions between the robotic agent and the environment and how this message can be decoded from the resulting motion trajectory. This research marks the beginning of a broader enquiry, with potential limitations to address and future directions to pursue. First, a formal secrecy evaluation through steganalysis is necessary to assess vulnerability to analytical detection mechanisms tailored to this domain. Second, the capacity of the steganographic system, measured in bits per trajectory, could be further optimised through more efficient coding methodologies. Third, factors such as cyber-physical gaps, observational imperfections and the presence of active adversaries could compromise steganographic communication, and therefore robustness against such errors at the receiving end warrants investigation. We envision that the concept of cyber-physical steganography in robotics will pave the way for broadening the scope of what constitutes channels for steganographic communication.

%\section*{Acknowledgements}
%This work was supported in part by the Japan Society for the Promotion of Science (JSPS) under KAKENHI Grants (JP21H04907 and JP24H00732) and the Japan Science and Technology Agency (JST) under CREST Grants (JPMJCR18A6 and JPMJCR20D3) and AIP Acceleration Grants (JPMJCR24U3).

\IEEEtriggeratref{34}
\bibliographystyle{Transactions-Bibliography/IEEEtran}
\bibliography{Transactions-Bibliography/bstcontrol, Bib/bib_armstego}

\begin{thebibliography}{10}
\providecommand{\url}[1]{#1}
\csname url@samestyle\endcsname
\providecommand{\newblock}{\relax}
\providecommand{\bibinfo}[2]{#2}
\providecommand{\BIBentrySTDinterwordspacing}{\spaceskip=0pt\relax}
\providecommand{\BIBentryALTinterwordstretchfactor}{4}
\providecommand{\BIBentryALTinterwordspacing}{\spaceskip=\fontdimen2\font plus
\BIBentryALTinterwordstretchfactor\fontdimen3\font minus \fontdimen4\font\relax}
\providecommand{\BIBforeignlanguage}[2]{{%
\expandafter\ifx\csname l@#1\endcsname\relax
\typeout{** WARNING: IEEEtran.bst: No hyphenation pattern has been}%
\typeout{** loaded for the language `#1'. Using the pattern for}%
\typeout{** the default language instead.}%
\else
\language=\csname l@#1\endcsname
\fi
#2}}
\providecommand{\BIBdecl}{\relax}
\BIBdecl

\bibitem{10.1007/3-540-61996-8_27}
D.~Kahn, ``The history of steganography,'' in \emph{Proc. Int. Workshop Inf. Hiding (IH)}, Cambridge, UK, 1996, pp. 1--5.

\bibitem{4655281}
N.~F. Johnson and S.~Jajodia, ``Exploring steganography: {S}eeing the unseen,'' \emph{Computer}, vol.~31, no.~2, pp. 26--34, 1998.

\bibitem{668971}
R.~Anderson and F.~Petitcolas, ``On the limits of steganography,'' \emph{IEEE J. Sel. Areas Commun.}, vol.~16, no.~4, pp. 474--481, 1998.

\bibitem{771065}
F.~Petitcolas, R.~Anderson, and M.~Kuhn, ``Information hiding{\textemdash}{A} survey,'' \emph{Proc. IEEE}, vol.~87, no.~7, pp. 1062--1078, 1999.

\bibitem{935180}
D.~Artz, ``Digital steganography: {H}iding data within data,'' \emph{IEEE Internet Comput.}, vol.~5, no.~3, pp. 75--80, 2001.

\bibitem{Fridrich:2009aa}
J.~Fridrich, \emph{Steganography in Digital Media: {P}rinciples, Algorithms, and Applications}.\hskip 1em plus 0.5em minus 0.4em\relax Cambridge, UK: Cambridge University Press, 2009.

\bibitem{10.1080/0161-119291866883}
P.~Wayner, ``Mimic functions,'' \emph{Cryptologia}, vol.~16, no.~3, pp. 193--214, 1992.

\bibitem{10.1007/3-540-61996-8_48}
D.~Gruhl, A.~Lu, and W.~Bender, ``Echo hiding,'' in \emph{Proc. Int. Workshop Inf. Hiding (IH)}, Cambridge, UK, 1996, pp. 295--315.

\bibitem{5387237}
W.~Bender, D.~Gruhl, N.~Morimoto, and A.~Lu, ``Techniques for data hiding,'' \emph{IBM Syst. J.}, vol.~35, no. 3\&4, pp. 313--336, 1996.

\bibitem{1511007}
J.~Fridrich, M.~Goljan, P.~Lisonek, and D.~Soukal, ``Writing on wet paper,'' \emph{IEEE Trans. Signal Process.}, vol.~53, no.~10, pp. 3923--3935, 2005.

\bibitem{chang-clark-2014-practical}
C.-Y. Chang and S.~Clark, ``Practical linguistic steganography using contextual synonym substitution and a novel vertex coding method,'' \emph{Comput. Linguist.}, vol.~40, no.~2, pp. 403--448, 2014.

\bibitem{Zhu:2018aa}
J.~Zhu, R.~Kaplan, J.~Johnson, and L.~Fei-Fei, ``{HiDDeN}: {H}iding data with deep networks,'' in \emph{Proc. Eur. Conf. Comput. Vis. (ECCV)}, Munich, Germany, 2018, pp. 682--697.

\bibitem{ziegler-etal-2019-neural}
Z.~Ziegler, Y.~Deng, and A.~Rush, ``Neural linguistic steganography,'' in \emph{Proc. Conf. Empir. Methods Nat. Lang. Process. (EMNLP)}, Hong Kong, China, 2019, pp. 1210--1215.

\bibitem{10.1007/10719724_5}
A.~Westfeld and A.~Pfitzmann, ``Attacks on steganographic systems,'' in \emph{Proc. Int. Workshop Inf. Hiding (IH)}, Dresden, Germany, 2000, pp. 61--76.

\bibitem{1040098}
H.~Farid, ``Detecting hidden messages using higher-order statistical models,'' in \emph{Proc. Int. Conf. Image Process.}, vol.~2, Rochester, NY, USA, 2002, pp. 905--908.

\bibitem{1203220}
N.~Provos and P.~Honeyman, ``Hide and seek: {A}n introduction to steganography,'' \emph{IEEE Secur. Priv.}, vol.~1, no.~3, pp. 32--44, 2003.

\bibitem{10.1145/1411328.1411349}
A.~D. Ker, T.~Pevn\'{y}, J.~Kodovsk\'{y}, and J.~Fridrich, ``The square root law of steganographic capacity,'' in \emph{Proc. ACM Workshop Multimed. Secur.}, Oxford, UK, 2008, pp. 107--116.

\bibitem{6081929}
J.~Kodovsk{\'y}, J.~Fridrich, and V.~Holub, ``Ensemble classifiers for steganalysis of digital media,'' \emph{IEEE Trans. Inf. Forensics Secur.}, vol.~7, no.~2, pp. 432--444, 2012.

\bibitem{6197267}
J.~Fridrich and J.~Kodovsk{\'y}, ``Rich models for steganalysis of digital images,'' \emph{IEEE Trans. Inf. Forensics Secur.}, vol.~7, no.~3, pp. 868--882, 2012.

\bibitem{7444146}
G.~Xu, H.~Wu, and Y.-Q. Shi, ``Structural design of convolutional neural networks for steganalysis,'' \emph{IEEE Signal Process. Lett.}, vol.~23, no.~5, pp. 708--712, 2016.

\bibitem{10.5555/96732}
T.~Yoshikawa, \emph{Foundations of robotics}.\hskip 1em plus 0.5em minus 0.4em\relax Cambridge, MA, USA: MIT Press, 1990.

\bibitem{10.1145/174147.174150}
B.~Donald, P.~Xavier, J.~Canny, and J.~Reif, ``Kinodynamic motion planning,'' \emph{J. ACM}, vol.~40, no.~5, pp. 1048--1066, 1993.

\bibitem{sutton1998reinforcement}
R.~Sutton and A.~Barto, \emph{Reinforcement Learning: {A}n Introduction}.\hskip 1em plus 0.5em minus 0.4em\relax Cambridge, MA, USA: MIT Press, 1998.

\bibitem{Mnih:2015aa}
V.~Mnih \emph{et~al.}, ``Human-level control through deep reinforcement learning,'' \emph{Nature}, vol. 518, no. 7540, pp. 529--533, 2015.

\bibitem{Collins:2024aa}
K.~M. Collins \emph{et~al.}, ``Building machines that learn and think with people,'' \emph{Nat. Hum. Behav.}, vol.~8, no.~10, pp. 1851--1863, 2024.

\bibitem{Floridi:2018aa}
L.~Floridi \emph{et~al.}, ``{AI4People}{\textemdash}{A}n ethical framework for a good {AI} society: {O}pportunities, risks, principles, and recommendations,'' \emph{Minds Mach.}, vol.~28, no.~4, pp. 689--707, 2018.

\bibitem{10.1145/3306618.3314289}
J.~Whittlestone, R.~Nyrup, A.~Alexandrova, and S.~Cave, ``The role and limits of principles in {AI} ethics: {T}owards a focus on tensions,'' in \emph{Proc. AAAI/ACM Conf. AI Ethics Soc. (AIES)}, Honolulu HI USA, 2019, pp. 195--200.

\bibitem{8662743}
A.~F. Winfield, K.~Michael, J.~Pitt, and V.~Evers, ``Machine ethics: {T}he design and governance of ethical {AI} and autonomous systems,'' \emph{Proc. IEEE}, vol. 107, no.~3, pp. 509--517, 2019.

\bibitem{10.5555/561828}
R.~M. Murray, S.~S. Sastry, and L.~Zexiang, \emph{A Mathematical Introduction to Robotic Manipulation}.\hskip 1em plus 0.5em minus 0.4em\relax Boca Raton, FL, USA: CRC Press, 1994.

\bibitem{7583659}
N.~Correll \emph{et~al.}, ``Analysis and observations from the first {A}mazon picking challenge,'' \emph{IEEE Trans. Autom. Sci. Eng.}, vol.~15, no.~1, pp. 172--188, 2018.

\bibitem{1389727}
N.~Koenig and A.~Howard, ``Design and use paradigms for {G}azebo, an open-source multi-robot simulator,'' in \emph{Proc. IEEE/RSJ Int. Conf. Intell. Robot. Syst. (IROS)}, vol.~3, Sendai, Japan, 2004, pp. 2149--2154.

\bibitem{6386109}
E.~Todorov, T.~Erez, and Y.~Tassa, ``{MuJoCo}: {A} physics engine for model-based control,'' in \emph{Proc. IEEE/RSJ Int. Conf. Intell. Robot. Syst. (IROS)}, Vilamoura-Algarve, Portugal, 2012, pp. 5026--5033.

\bibitem{6696520}
E.~Rohmer, S.~P.~N. Singh, and M.~Freese, ``{V-REP}: {A} versatile and scalable robot simulation framework,'' in \emph{Proc. IEEE/RSJ Int. Conf. Intell. Robot. Syst. (IROS)}, Tokyo, Japan, 2013, pp. 1321--1326.

\bibitem{makoviychuk2021isaac}
V.~Makoviychuk \emph{et~al.}, ``{Isaac Gym}: {H}igh performance {GPU} based physics simulation for robot learning,'' in \emph{Proc. Int. Conf. Neural Inf. Process. Syst. (NeurIPS)}, Virtual Event, 2021, pp. 1--12.

\bibitem{1014739}
A.~Ijspeert, J.~Nakanishi, and S.~Schaal, ``Movement imitation with nonlinear dynamical systems in humanoid robots,'' in \emph{Proc. IEEE Int. Conf. Robot. Autom. (ICRA)}, vol.~2, Washington, DC, USA, 2002, pp. 1398--1403.

\bibitem{BELLMAN:1957aa}
R.~Bellman, ``A {M}arkovian decision process,'' \emph{J. Math. Mech.}, vol.~6, no.~5, pp. 679--684, 1957.

\bibitem{Sutton:1988aa}
R.~S. Sutton, ``Learning to predict by the methods of temporal differences,'' \emph{Mach. Learn.}, vol.~3, no.~1, pp. 9--44, 1988.

\bibitem{watkins1989learning}
C.~J. C.~H. Watkins, ``Learning from delayed rewards,'' Ph.D. dissertation, King's College, University of Cambridge, Cambridge, UK, 1989.

\bibitem{Lin:1992aa}
L.-J. Lin, ``Self-improving reactive agents based on reinforcement learning, planning and teaching,'' \emph{Mach. Learn.}, vol.~8, no.~3, pp. 293--321, 1992.

\bibitem{Williams:1992aa}
R.~J. Williams, ``Simple statistical gradient-following algorithms for connectionist reinforcement learning,'' \emph{Mach. Learn.}, vol.~8, no.~3, pp. 229--256, 1992.

\bibitem{NIPS1999_464d828b}
R.~S. Sutton, D.~McAllester, S.~Singh, and Y.~Mansour, ``Policy gradient methods for reinforcement learning with function approximation,'' in \emph{Proc. Int. Conf. Neural Inf. Process. Syst. (NeurIPS)}, vol.~12, Denver, CO, USA, 1999, pp. 1057--1063.

\bibitem{10.5555/3104482.3104569}
J.~Ngiam, A.~Khosla, M.~Kim, J.~Nam, H.~Lee, and A.~Y. Ng, ``Multimodal deep learning,'' in \emph{Proc. Int. Conf. Mach. Learn. (ICML)}, Bellevue, WA, USA, 2011, pp. 689--696.

\bibitem{JMLR:v15:srivastava14b}
N.~Srivastava and R.~Salakhutdinov, ``Multimodal learning with deep {B}oltzmann machines,'' \emph{J. Mach. Learn. Res.}, vol.~15, no.~84, pp. 2949--2980, 2014.

\bibitem{8269806}
T.~Baltru{\v s}aitis, C.~Ahuja, and L.-P. Morency, ``Multimodal machine learning: {A} survey and taxonomy,'' \emph{IEEE Trans. Pattern Anal. Mach. Intell.}, vol.~41, no.~2, pp. 423--443, 2019.

\bibitem{10.1145/3656580}
P.~P. Liang, A.~Zadeh, and L.-P. Morency, ``Foundations \& trends in multimodal machine learning: {P}rinciples, challenges, and open questions,'' \emph{ACM Comput. Surv.}, vol.~56, no.~10, pp. 1--42, 2024.

\bibitem{1087068}
O.~Khatib, ``A unified approach for motion and force control of robot manipulators: {T}he operational space formulation,'' \emph{IEEE J. Robot. Autom.}, vol.~3, no.~1, pp. 43--53, 1987.

\bibitem{6769090}
C.~E. Shannon, ``Communication theory of secrecy systems,'' \emph{Bell Syst. Tech. J.}, vol.~28, no.~4, pp. 656--715, 1949.

\bibitem{li2024evaluating}
X.~Li \emph{et~al.}, ``Evaluating real-world robot manipulation policies in simulation,'' in \emph{Proc. Conf. Robot Learn. (CoRL)}, Munich, Germany, 2024, pp. 1--24.

\bibitem{kim2024openvla}
M.~J. Kim \emph{et~al.}, ``{OpenVLA}: {A}n open-source vision-language-action model,'' in \emph{Proc. Conf. Robot Learn. (CoRL)}, Munich, Germany, 2024, pp. 1--35.

\bibitem{Ghosh-RSS-24}
D.~Ghosh \emph{et~al.}, ``{Octo}: {A}n open-source generalist robot policy,'' in \emph{Proc. Robot. Sci. Syst. (RSS)}, Delft, Netherlands, 2024, pp. 1--13.

\end{thebibliography}

\vfill

\end{document}